\documentclass[10pt,letterpaper,twocolumn,10pt,letterpaper,twocolumn]{article}
\usepackage[latin9]{inputenc}
\usepackage{color}
\usepackage{array}
\usepackage{multirow}
\usepackage{amsmath}
\usepackage{amssymb}
\usepackage{graphicx}
\usepackage[unicode=true,
 bookmarks=false,
 breaklinks=true,pdfborder={0 0 1},backref=section,colorlinks=false]
 {hyperref}
\hypersetup{
 pagebackref=true,letterpaper=true,colorlinks}

\makeatletter

\pdfpageheight\paperheight
\pdfpagewidth\paperwidth

\providecommand{\tabularnewline}{\\}

\usepackage{cvpr}
\usepackage{times}
\usepackage{epsfig}

 \cvprfinalcopy % *** Uncomment this line for the final submission

\ifcvprfinal\fi

\makeatother

\begin{document}

\title{Pose-Normalized Image Generation for Person Re-identification}

\author{Xuelin Qian$^{1}$, Yanwei Fu$^{1}$, Tao Xiang$^{2}$,
Wenxuan Wang$^{1}$ \\
Jie Qiu$^{3}$, Yang Wu$^{3}$, Yu-Gang Jiang$^{1}$,
Xiangyang Xue$^{1}$\\
$^{1}$Fudan University; $^{2}$Queen Mary University
of London; \\
$^{3}$Nara Institute of Science and Technology; \\
}
\maketitle
\begin{abstract}
Person Re-identification (re-id) faces two major challenges: the lack
of cross-view paired training data and learning discriminative identity-sensitive
and view-invariant features in the presence of large pose variations.
In this work, we address both problems by proposing a novel deep person
image generation model for synthesizing realistic person images conditional
on pose. The model is based on a generative adversarial network (GAN)
designed specifically for pose normalization in re-id, thus termed
pose-normalization GAN (PN-GAN). With the synthesized images, we can
learn a new type of deep re-id feature free of the influence of pose
variations. We show that this feature is strong on its own and complementary
to features learned with the original images. Importantly, under the
transfer learning setting, we show that our model generalizes well to
any new re-id dataset without the need for collecting any training
data for model fine-tuning. The model thus has the potential to make  re-id
model truly scalable. 
\end{abstract}

\section{Introduction}

\begin{figure}[t]
\begin{centering}
\includegraphics[scale=0.3]{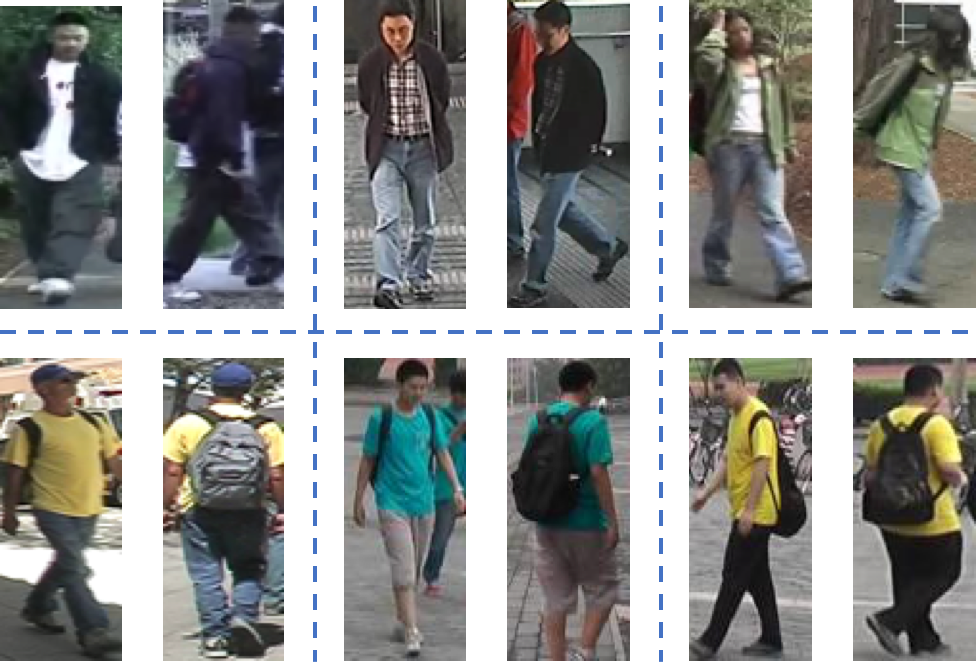} 
\par\end{centering}
\caption{The same person's appearance can be very different across camera views,
due to the presence of large pose variations.}
\label{fig:The-different_pose} 
\end{figure}

Person Re-identification (re-id) aims to match a person across multiple
non-overlapping camera views \cite{gong2011person}. It is a very
challenging problem because a person's appearance can change drastically
across views, due to the changes in various covariate factors independent
of the person's identity. These factors include viewpoint, body configuration,
lighting, and occlusion (see Fig.~\ref{fig:The-different_pose}).
Among these factors, pose plays the most important role in causing
a person's appearance changes. Here pose is defined as a combination
of viewpoint and body configuration. It is thus also a cause of self-occlusion.
For instance, in the bottom row examples in Fig.~\ref{fig:The-different_pose},
the big backpacks carried by the three persons are in full display
from the back, but reduced to mostly the straps from the front.

Most existing re-id approaches \cite{deepreid,Ejaz_cvpr2015,de_cheng_2016,yu2017cross,qian2017multi,SVDNet,reid_in_wild,PersonSearch}
are based on learning identity-sensitive and view-insensitive features
using deep neural networks (DNNs). To learn the features, a large
number of persons' images need to be collected in each camera view
with variable poses. With the collected images, the model can have a chance to learn what features
are discriminative and invariant to the camera view and pose changes.
These approaches thus have a number of limitations. The first limitation
is \textbf{lack of scalability} to large camera networks. Existing
models require sufficient identities and sufficient images per identity
to be collected from each camera view. However, manually annotating
persons across views in the camera networks is tedious and difficult
even for humans. Importantly, in a real-world application, a camera
network can easily consist of hundreds of cameras (i.e. those in an
airport or shopping mall); annotating enough training identities from
all camera views are infeasible. The second limitation is \textbf{lack
of generalizability} to new camera networks. Specifically, when an
existing deep re-id model is deployed to a new camera network, view
points and body poses are often different across the networks; additional data thus need
to be collected for model fine-tuning, which severely limits its generalization
ability. As a result of both limitations, although deep re-id models
are far superior for large re-id benchmarks such as Market-1501 \cite{market1501}
and CUHK03 \cite{deepreid}, they still struggle to beat hand-crafted
feature based models on smaller datasets such as CUHK01 \cite{cuhk01},
even when they are pre-trained on the larger re-id datasets.

Even with sufficient labeled training data, existing deep re-id models
face the challenge of learning identity-sensitive and view-insensitive
features in the presence of large pose variations. This is because
a person's appearance is determined by a combination of identity-sensitive
but view-insensitive factors and identity-insensitive but view-sensitive
ones, which are inter-connected. The former correspond to semantic
related identity properties, such as gender, carrying, clothing style,
color, and texture. The latter are the covariates mentioned earlier
including pose. Existing models aim to keep the former and remove
the latter in the learned feature representations. However, these
two aspects of the appearance are not independent, e.g., the appearance
of the carrying depends on the pose. Making the learned features pose-insensitive
means that the features supposed to represent the backpacks in the
bottom row examples in Fig.~\ref{fig:The-different_pose} are reduced
to those representing only the straps \textendash{} a much harder
type of features to learn.

In this paper, we argue that the key to learning an effective, scalable
and generalizable re-id model is to remove the influence of pose on
the person's appearance. Without the pose variation, we can learn
a model with much less data thus making the model scalable to large
camera networks. Furthermore, without the need to worry about the
pose variation, the model can concentrate on learning identity-sensitive
features and coping with other covariates such as different lighting
conditions and backgrounds. The model is thus far more likely to generalize
to a new dataset from a new camera network. Moreover, with the different
focus, the features learned without the presence of pose variation
would be different and complementary to those learned with pose variation.

To this end, a novel deep re-id framework is proposed. Key to the
framework is a deep person image generation model. The model is based
on a generative adversarial network (GAN) designed specifically for
pose normalization in re-id. It is thus termed pose-normalization GAN (PN-GAN).
Given any person's image and a desirable pose as input, the model
will output a synthesized image of the same identity with the original
pose replaced with the new one. In practice, we define a set of eight
canonical poses, and synthesize eight new images for any given image,
resulting in a 8-fold increase in the training data size. The pose-normalized
images are used to train a pose-normalized re-id model which produces
a set of features that are complementary to the feature learned with
the original images. The two sets of feature are thus fused as the final feature representation.
Critically, once trained, the model can be applied to a new dataset
without any model fine-tuning as long as the test image's pose is
also normalized.

\noindent \textbf{Contributions}. Our contributions are as follows.
(1) We identify pose as the chief culprit for preventing a deep re-id
model from learning effective identity-sensitive and view-insensitive
features, and propose a novel solution based on generating pose-normalized
images. This also addresses the scalability and generalizability issues
of existing models. (2) A novel person image generation model PN-GAN
is proposed to generate pose-normalized images, which are realistic,
identity-preserving and pose controllable. With the synthesized images
of canonical poses, strong and complementary features are learned
to be combined with features learned with the original images. Extensively
experiments on several benchmarks show that the efficacy of our proposed
model. (3) A more realistic unsupervised transfer learning setting is considered
in this paper. Under this setting, no data from the target dataset
is used for model updating: the model trained from labeled source
datasets/domains is applied to the target domain without any modification.

\section{Related Work}

\noindent \textbf{Deep re-id models}\quad{}Most recently proposed
re-id models employ a DNN to learn discriminative view-invariant features
\cite{deepreid,Ejaz_cvpr2015,de_cheng_2016,yu2017cross,qian2017multi,SVDNet,reid_in_wild,PersonSearch}.
They differ in the DNN architectures \textendash{} some adopt a standard
DNN developed for other tasks, whilst others have architectures tailor-made.
They differ also in the training objectives. Different models use different
training losses including identity classification, pairwise verification,
and triplet ranking losses. A comprehensive study on the effectiveness
of different losses and their combinations on re-id can be found in
\cite{deeptransfer2016}. The focus of this paper is not on designing
new re-id deep model architecture or loss \textendash{} we use an
off-the-shelf ResNet architecture \cite{resnet} and the standard
identity classification loss. We show that once the pose variation
problem is solved, such a general-purpose model can achieve the state-of-the-art
re-id performance, beating many existing models with more elaborative
architectures and losses.

\noindent \textbf{Pose-guided deep re-id} \quad{}The negative effects
of pose variation on deep re-id models have been recognised recently.
A number of models \cite{su2017pose,zheng2017pose,zhao2017deeply,zhao2017spindle,li2017learning,yao2017deep,wei2017glad}
are proposed to address this problem. Most of them are pose-guided
based on body part detection. For example, \cite{su2017pose,zhao2017spindle}
utilize detect normalized part regions from a person image, and then
fuse the features extracted from the original images and the part
region images. These body part regions are predefined and the region
detectors are trained beforehand. Differently, \cite{zhao2017deeply}
combine region selection and detection with deep re-id in one model.
Our model differs significantly from these models in that we synthesize
realistic whole-body images using the proposed PN-GAN, rather than
only focusing on body parts for pose normalization. Note that body
parts are related to semantic attributes which are often specific to
different body parts. A number of attributes based re-id models \cite{wang2017attribute,sarfraz2017deep,yu2016weakly,deng2015learning}
have been proposed. They use attributes to provide additional supervision
for learning identity-sensitive features. In contrast, without using
the additional attribute information,  our PN-GAN is learned 
as a conditional image generation model for the re-id problem. 

\noindent \textbf{Deep image generation} \quad{}Generating realistic
images of objects using DNNs has received much interest recently,
thanks largely to the development of GAN~\cite{goodfellow2014generative}.
GAN is designed to find the optimal discriminator network $D$ between
training data and generated samples using a min-max game and simultaneously
enhance the performance of an image generator network $G$. It is
formulated to optimize the following objective functions: 
\begin{align}
\underset{G}{\mathrm{min}}\underset{D}{\mathrm{max}}\mathcal{L}_{GAN} & =\mathbb{E}_{x\sim p_{data}\left(x\right)}\left[\mathrm{log}D\left(x\right)\right]+\label{eq:gan}\\
 & \mathbb{E}_{z\sim p_{prior}\left(z\right)}\left[\mathrm{log}\left(1-D\left(G\left(z\right)\right)\right)\right]\nonumber 
\end{align}
where $p_{data}\left(x\right)$ and $p_{prior}\left(z\right)$ are
the distributions of real data $x$ and Gaussian prior $z\sim\mathcal{N}\left(\mathbf{0},\mathbf{1}\right)$.
The training process iteratively updates the parameters of $G$ and
$D$ with the loss functions $\mathcal{L}_{D}=-\mathcal{L}_{GAN}$
and $\mathcal{L}_{G}=\mathcal{L}_{GAN}$ for the generator and discriminator
respectively. The generator can draw a sample $z\sim p_{prior}\left(z\right)=\mathcal{N}\left(\mathbf{0},\mathbf{1}\right)$
and utilize the generator network $G$, i.e., $G(z)$ to generate
an image.

\noindent Among all the variants of GAN, our pose normalization GAN
is built upon deep convolutional generative adversarial networks (DCGANs)
\cite{radford2015unsupervised}.
Based on a standard convolutional decoder, DCGAN scales up GAN using
Convolutional Neural Networks (CNNs) and it results in stable training
across various datasets. Many other variants of GAN, such as VAEGAN
\cite{larsen2015autoencoding}, Conditional GAN \cite{isola2016image},
stackGAN \cite{zhang2016stackgan} also exist. However, most of
them are designed for training with high-quality images of objects
such as celebrity faces, instead of low-quality surveillance video
frames of pedestrians. This problem is tackled in a very recent work
\cite{poseguid2017nips}, which also aims to synthesize person images
in different poses.

Nonetheless, our model differs significant from the existing variants of GAN. In particular,  built upon
the residual blocks, our PN-GAN is learned to change the poses and
yet keeps the identity of input person. Note that the only work so
far that uses deep image generator for re-id is \cite{zheng2017unlabeled}.
However, their model is not a conditional GAN and thus cannot
control either identity or pose in the generated person images. As
a result, the generated images can only be used as unlabeled or weakly
labeled data. In contrast, our model generate strongly labeled data
with its ability to preserve the identity and remove the influence
of pose variation.

\begin{figure}
\begin{centering}
\includegraphics[scale=0.1]{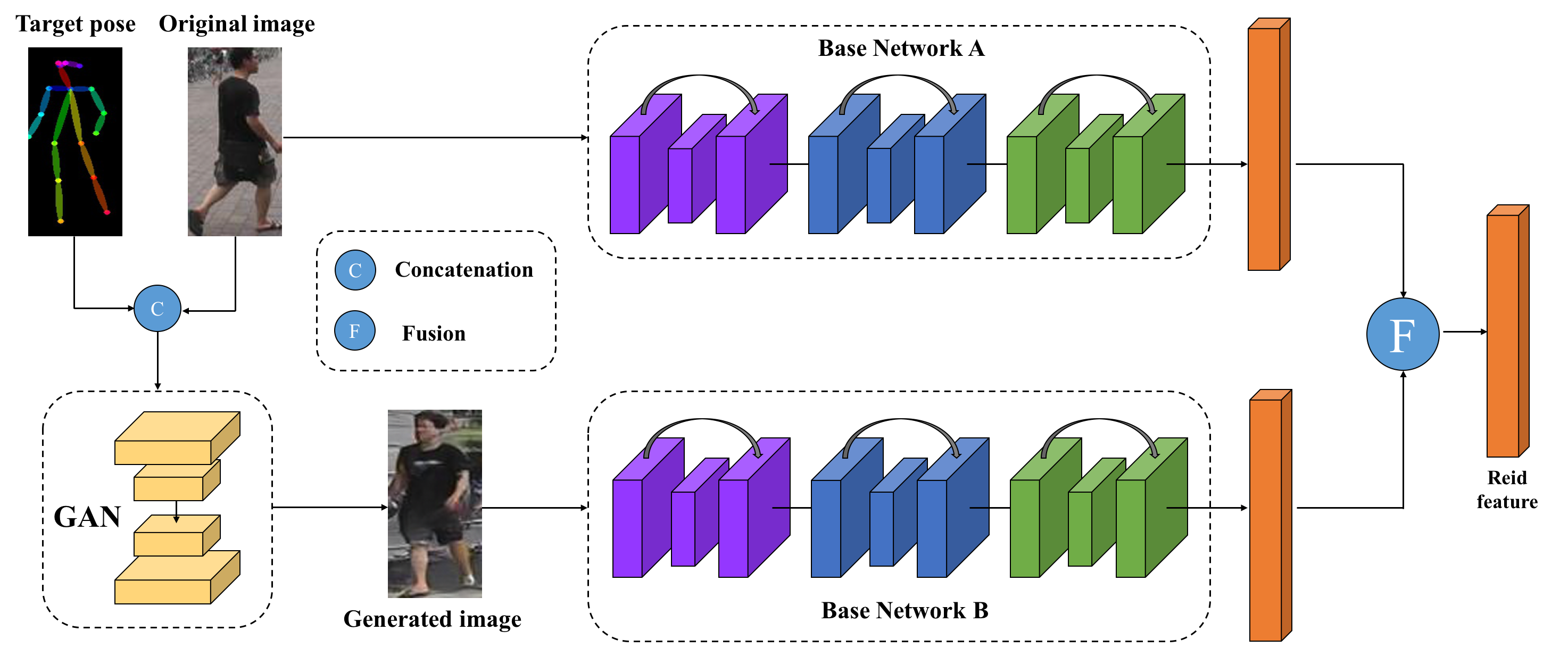}
\par\end{centering}
\caption{\label{fig:Overview} Overview of our framework.}
\end{figure}

\vspace{-0.5cm}
\section{Methodology}

\subsection{Problem Definition and Overview\label{subsec:Problem-Setup-and}}

\noindent \textbf{Problem definition}. Assume we have a training dataset
of $N$ persons $\mathcal{D}_{Tr}=\left\{ \mathbf{I}_{k},y_{k}\right\} _{k=1}^{N}$,
where $\mathbf{I}_{k}$ and $y_{k}$ are the person image and person
id of the $k$-th person. In the training stage we learn a feature
extraction function $\phi$ so that a given image $\mathbf{I}$ can
be represented by a feature vector $\mathbf{f}_{\mathbf{I}}=\phi(\mathbf{I})$.
In the testing stage, given a pair of person images $\left\{ \mathbf{I}_{i},\mathbf{I}_{j}\right\} $
in the testing dataset $\mathcal{D}_{Te}$, we need to judge whether
$y_{i}=y_{j}$ or $y_{i}\neq y_{j}$. This is done by simply computing
the Euclidean distance between $\mathbf{f}_{\mathbf{I}_{i}}$ and
$\mathbf{f}_{\mathbf{I}_{j}}$ as the identity-similarity measure.

\noindent \textbf{Framework Overview}. As shown in Fig.~\ref{fig:Overview},
our framework has two key components, \emph{i.e.}, a GAN based person
image generation model (Sec.~\ref{subsec:GAN-based-person}) and
a person re-id feature learning model (Sec.~\ref{subsec:person-re-id-classification}).

\subsection{Deep Image Generator\label{subsec:GAN-based-person}}

\noindent Our image generator aims at producing the same person's
images under different poses. Particularly, given an input person
image $\mathbf{I}_{i}$ and a desired pose image $\mathbf{I}_{\mathcal{P}_{j}}$,
our image generator aims to synthesize a new person image $\hat{\mathbf{I}}_{j}$,
which contains the same person but with a different pose defined by
$\mathbf{I}_{\mathcal{P}_{j}}$. As in any GAN model, the image generator
has two components, a Generator $G_{P}$ and a Discriminator $D_{P}$.
The generator is learned to edit the person image conditional on a
given pose; the discriminator discriminates real data samples from
the generated samples and help to improve the quality of generated
images.

\begin{figure}
\begin{centering}
\includegraphics[scale=0.1]{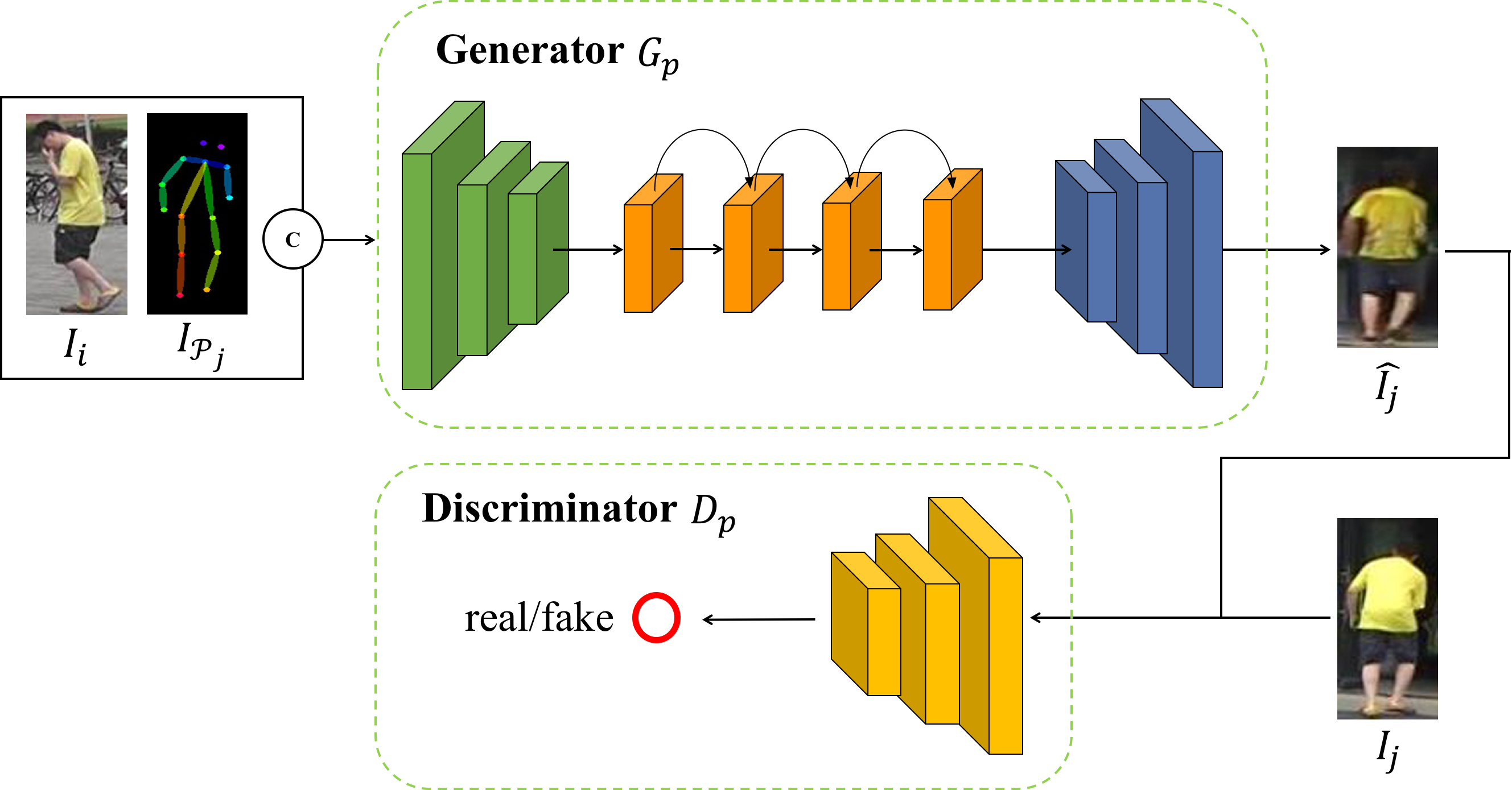}
\par\end{centering}
\caption{\label{fig:gan} Schematic of our PN-GAN model}
\end{figure}

\vspace{0.1in}

\noindent \textbf{Pose estimation.} The image generation process is
conditional on the input image and one  factor: the desired pose represented
by a skeleton pose image. Pose estimation is obtained by a pretrained
off-the-shelf model. More concretely, the off-the-shelf pose detection
toolkit \textendash{} OpenPose \cite{cao2017realtime} is deployed,
which is trained without using any re-id benchmark data. Given an
input person image $\mathbf{I}_{i}$, the pose estimator can produce
a pose image $\mathbf{I}_{\mathcal{P}_{i}}$, which localizes and
detects 18 anatomical key-points as well as their connections. In
the pose images, the orientation of limbs is encoded by color (see
Fig.~\ref{fig:Overview}, target pose). In theory, any pose from
any person image can be used as a condition to control the pose of
another person's generated image. In this work, we focus on pose normalization
so we stick to eight canonical poses as shown in Fig.~\ref{fig:The-eight-poses}(a),
to be detailed later.

\vspace{0.1in}

\noindent \textbf{Generator.} As shown in Fig.~\ref{fig:gan}, given
an input person image $\mathbf{I}_{i}$, and a target person image
$\mathbf{I}_{j}$ which contains the same person as $\mathbf{I}_{i}$
but a different pose $\mathbf{I}_{\mathcal{P}_{j}}$, our generator
will learn to replace pose information in $\mathbf{I}_{i}$ with the
target pose $\mathbf{I}_{\mathcal{P}_{j}}$ and generate the new pose
$\hat{\mathbf{I}}_{j}$. The input to the generator is the concatenation
of the input person image $\mathbf{I}_{i}$ and target pose image
$\mathbf{I}_{\mathcal{P}_{j}}$. Specifically, we treat the target
body pose image $\mathbf{I}_{\mathcal{P}_{j}}$ as a three-channel
image and directly concatenate it with the three-channel source person
image as the input of the generator. The generator $G_{P}$ is designed
based on the ``ResNet'' architecture and is an encoder-decoder network
\cite{hinton2006science}. The encoder-decoder network progressively
down-samples $\mathbf{I}_{i}$ to a bottleneck layer, and then reverse
the process to generate $\hat{\mathbf{I}}_{j}$. The encoder contains $9$ ResNet basic blocks\footnote{Details of structure are in the Supplementary.}. 

The motivation of designing such a generator is to take advantage
of learning residual information in generating new images. The general
shape of ``ResNet'' is learning $y=f(x)+x$ which can be used to
pass invariable information from the bottom layers of the encoder to the decoder,
and change the variable information of pose. To this end, the other
features (e.g., clothing, and the background) will also be reserved
and passed to the decoder in order to generate $\hat{\mathbf{I}}_{j}$. With
this architecture (see Fig.~\ref{fig:gan}), we have the best of
both worlds: the encoder-decoder network can help learn to extract
the semantic information, stored in the bottleneck layer, while  the ResNet
blocks can pass rich invariable information of person identity to
help synthesize more realistic images, and change variable information
of poses to realize pose normalization at the same time.

Formally, let $G_{P}\left(\cdot\right)$ be the generator network
which is composed of an encoder subnet $G_{Enc}\left(\cdot\right)$
and a decoder subnet $G_{Dec}\left(\cdot\right)$, the objective of
the generator network can be expressed as 
\begin{equation}
\mathcal{L}_{_{G_{P}}=}\mathcal{L}_{GAN}+\lambda_{1}\cdot\mathcal{L}_{L_{1}},\label{eq:generator}
\end{equation}
where $\mathcal{L}_{GAN}$ is the loss of the generator in Eq (\ref{eq:gan})
with the generator $G_{P}\left(\cdot\right)$ and discriminator $D_{P}\left(\cdot\right)$
respectively, 
\begin{align}
\mathcal{L}_{GAN} & =\mathbb{E}_{\mathbf{I}_{j}\sim p_{data}\left(\mathbf{I}_{j}\right)}\left\{ \mathrm{log}D_{P}\left(\mathbf{I}_{j}\right)\right.\label{eq:updated_GAN}\\
+ & \left.\mathrm{log}\left(1-D_{P}\left(G_{P}\left(\mathbf{I}_{i},\mathbf{I}_{\mathcal{P}_{j}}\right)\right)\right)\right\} \nonumber 
\end{align}

\noindent and $\mathcal{L}_{L_{1}}=\mathbb{E}_{\mathbf{I}_{j}\sim p_{data}\left(\mathbf{I}_{j}\right)}\left[\left\Vert \mathbf{I}_{j}-\hat{\mathbf{I}}_{j}\right\Vert _{1}\right]$,
and $\hat{\mathbf{I}}_{j}=G_{Dec}\left(G_{Enc}\left(\mathbf{I}_{i},\mathbf{I}_{\mathcal{P}_{j}}\right)\right)$
is the reconstructed image for $\mathbf{I}_{j}$ from the input image
$\mathbf{I}_{i}$ with the body pose $\mathbf{I}_{\mathcal{P}_{j}}$.
Here the $L_{1}-$norm is used to yield sharper and cleaner images.
$\lambda_{1}$ is the weighting coefficient to balance the importance
of each term.

\vspace{0.03in}

\noindent \textbf{Discriminator}. The discriminator $D_{P}\left(\cdot\right)$
aims at learning to differentiate the input images is real or fake
(i.e., a binary classification task). Given the input image $\mathbf{I}_{i}$
and target output image $\mathbf{I}_{j}$, the objective of the discriminator
network can be formulated as 
\begin{equation}
\mathcal{L}_{D_{P}}=-\mathcal{L}_{GAN},\label{eq:discriminator}
\end{equation}
Since our final goal is to obtain the best generator $G_{P}$, the
optimization step would be to iteratively minimize the loss function
$\mathcal{L}_{G_{P}}$ and $\mathcal{L}_{D_{P}}$ until convergence.
Please refer to the Supplementary Material for the detailed structures
and parameters of the generator and discriminator.

\subsection{Person re-id with Pose Normalization \label{subsec:person-re-id-classification}}

\begin{figure*}
\centering{}%
\begin{tabular}{cc}
\includegraphics[scale=0.15]{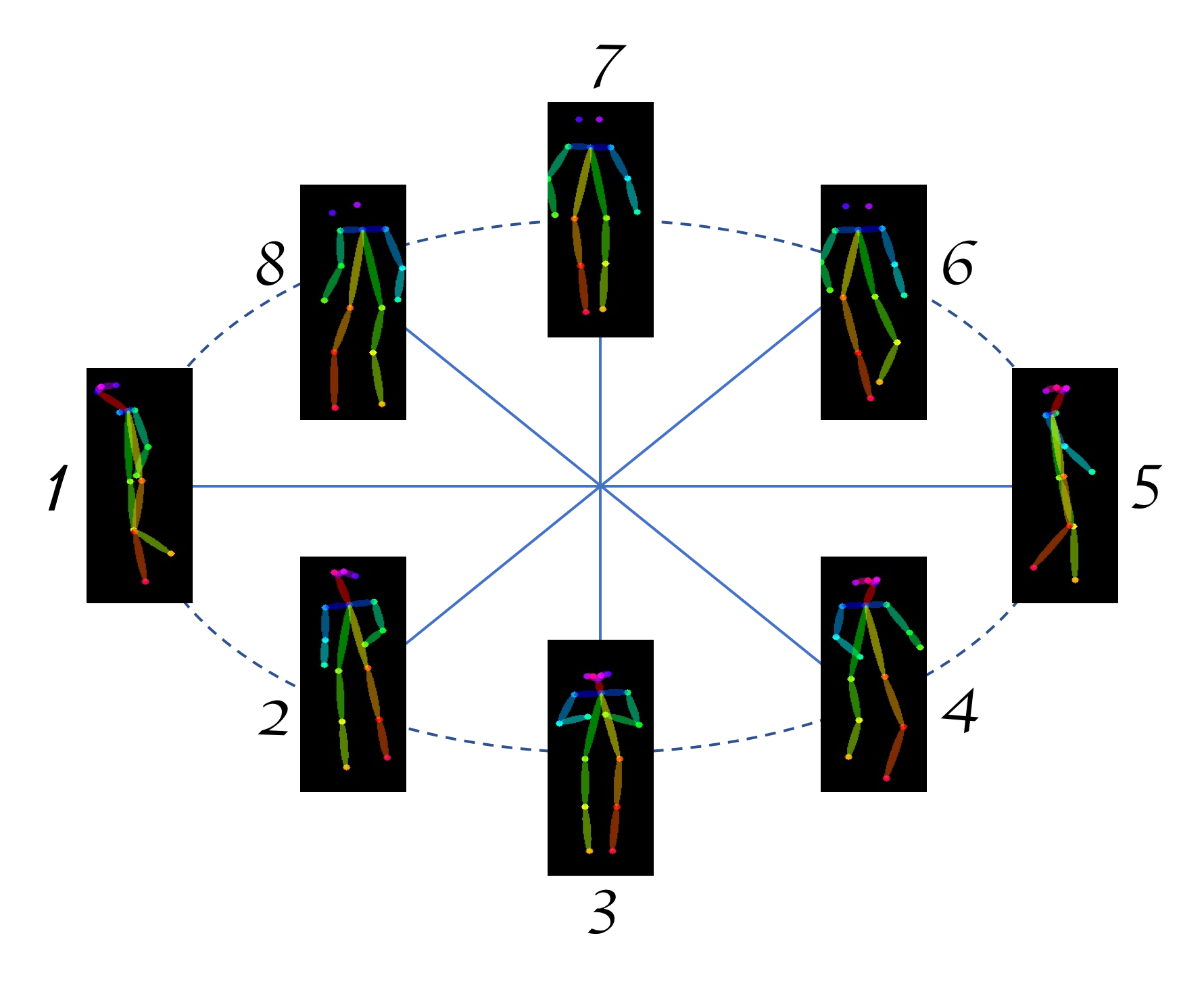}  & \includegraphics[scale=0.15]{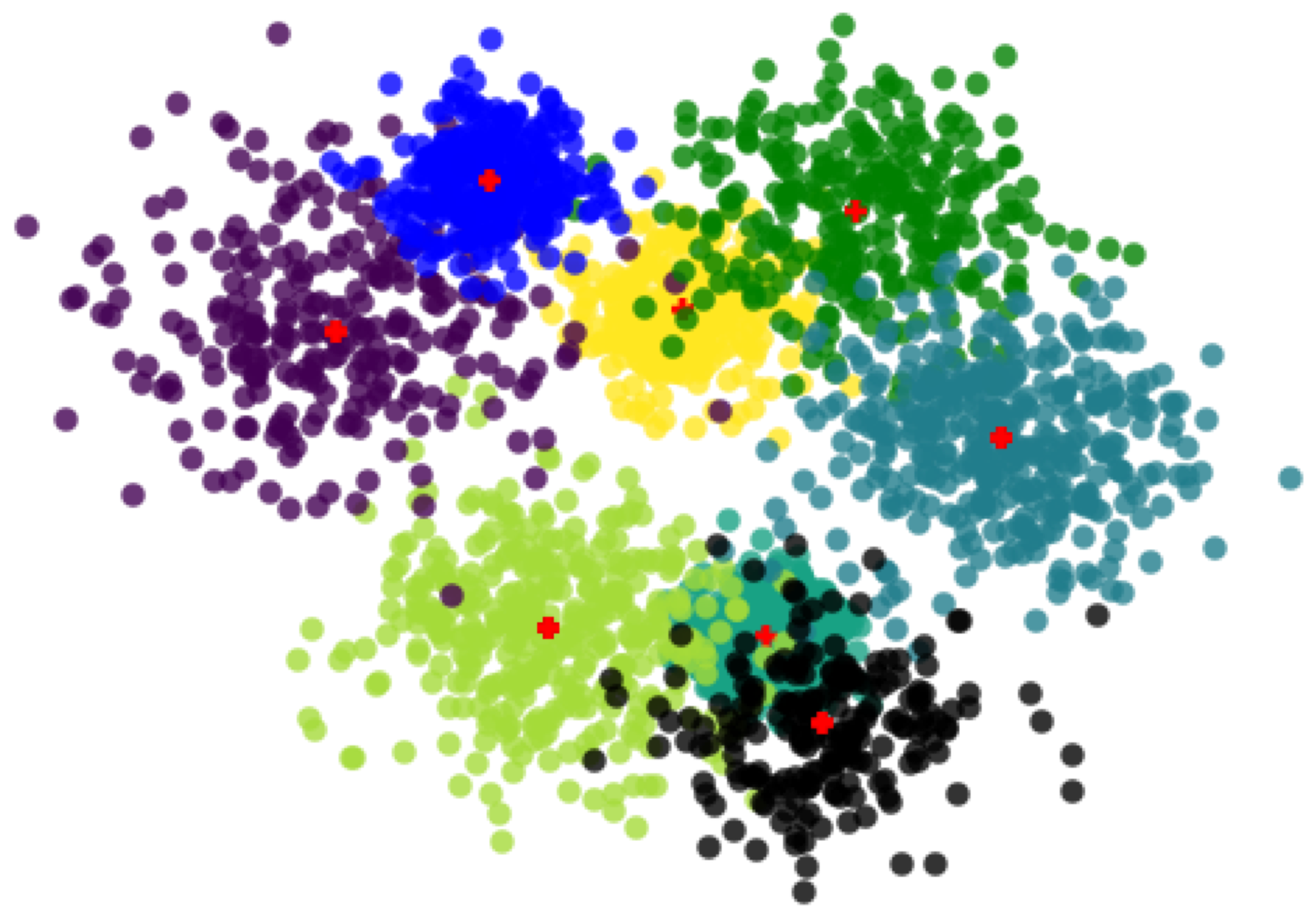}\tabularnewline
(a) Eight canonical poses on Market-1501  & (b) t-SNE visualization of different poses.\tabularnewline
\end{tabular}\caption{\label{fig:The-eight-poses} Visualization of canonical poses. Note
that red crosses in (b) indicates the canonical pose obtained as the cluster means.}
\end{figure*}

As shown in Fig.~\ref{fig:Overview}, we train two re-id models.
One model is trained using the original images in a training set to
extract identity-invariant features in the presence of pose variation.
The other is trained using the synthesized images with normalized
poses using our PN-GAN to compute re-id features free of pose variation.
They are then fused as the final feature representation.

\vspace{0.03in}

\noindent \textbf{Pose Normalization. } We need to obtain a set of
canonical poses, which are representative of the typical viewpoint
and body-configurations exhibited by people in public captured by
surveillance cameras. To this end, we predict the poses of all training
images in a dataset and then group the poses into eight clusters $\left\{ \mathbf{I}_{\mathcal{P}_{C}}\right\} _{c=1}^{8}$.
We use VGG-19 \cite{returnDevil2014BMVC} pre-trained on the ImageNet ILSVRC-2012
dataset to extract the features of each pose images, and K-means algorithm
is used to cluster the  training pose images into canonical poses.
The mean pose images of these clusters are then used as the canonical
poses. The eight poses obtained on Market-1501 \cite{market1501} is shown in Fig.~\ref{fig:The-eight-poses}(a).
With these poses, given each image $\mathbf{I}_{i}$, our generator
will synthesize eight images $\left\{ \hat{\mathbf{I}}_{i,\mathcal{P}_{C}}\right\} _{C=1}^{8}$
by replacing the original pose with these poses.

\vspace{0.03in}

\noindent \textbf{Re-id Feature with pose variation. } We train one
re-id model with the original training images to extract re-id features with pose variation. The ResNet-50 model \cite{resnet}
is used as the base network. It is pre-trained on the
ILSVRC-2012 dataset, and fine-tuned on the training set of a given
re-id dataset to classify the training identities. We name this network
ResNet-50-A (Base Network A), as shown in Fig.~(\ref{fig:Overview}). Given
an input image $\mathbf{I}_{i}$, ResNet-50-A produces a feature set
$\left\{ \mathbf{f}_{\mathbf{I}_{i},layer}\right\} $, where $layer$
indicates from which layer of the network, the re-id features are
extracted. Note that, in most existing deep re-id models, features
are computed from the final convolutional layer. Inspired by \cite{liu2017hydraplus}
which shows that layers before the final layer in a DNN often contain
useful mid-level identity-sensitive information. We thus merge the
$5a$, $5b$ and $5c$ convolutional layers of ResNet-50 structures
into a $1024\textendash{}d$ feature vector after an FC layer.

\noindent \textbf{Re-id Feature without pose variation. } The second
model called ResNet-50-B has the same architecture as ResNet-50-A, but performs
feature learning using the pose-normalized synthetic images. We thus
obtain eight sets of features for the eight poses $\mathbf{f}_{\hat{\mathbf{I}}{}_{i,\mathcal{P}_{C}}}=\left\{ \mathbf{f}_{\hat{\mathbf{I}}_{i,\mathcal{P}_{C}}}\right\} _{C=1}^{8}$.

\noindent \textbf{Testing stage. } Once ResNet-50-A and ResNet-50-B
are trained, during testing, for each gallery image, we feed it into
ResNet-50-A to obtain one feature vector; and synthesize eight images
of the canonical poses, feed them into ResNet-50-B to obtain 8 pose-free
features. This can be done offline. Then given a query image $\mathbf{I}_{q}$,
we do the same to obtain nine feature vectors $\left\{ \mathbf{f}_{\mathbf{I}_{q}},\mathbf{f}_{\hat{\mathbf{I}}_{q,\mathcal{P}_{C}}}\right\} $.
Since Maxout and Max-pooling have been widely used in multi-query
video re-id, we thus obtain one final feature vector by fusing the
nine feature vectors by element-wise maximum operation. We then calculate
the Euclidean distance between the final feature vectors of the query
and gallery images and use the distance to rank the gallery images.

\section{Experiments}

\subsection{Datasets and Settings}

\noindent Experiments are carried out on four benchmark datasets: 

\noindent \noindent \textbf{Market-1501} \cite{market1501} is collected from 6
different camera views. It has 32,668 bounding boxes of 1,501 identities
obtained using a Deformable Part Model (DPM) person detector. Following
the standards split \cite{market1501}, we use 751 identities with
12,936 images as training and the rest 750 identities with 19,732
images for testing. The training set is used to train our PN-GAN model.
\textbf{ }

\noindent \noindent \textbf{CUHK03} \cite{deepreid} contains 14,096 images of
1,467 identities, captured by six camera views with 4.8 images for
each identity in each camera on average. We utilize the more realistic
yet harder detected person images setting. The training, validation
and testing sets consist of 1,367 identities, 100 identities and 100
identities respectively. The testing process is repeated with 20 random
splits following \cite{deepreid}. \textbf{ }

\noindent \noindent \textbf{DukeMTMC-reID} \cite{Duke_ori_data} is  constructed from the
multi-camera tracking dataset \textendash{} DukeMTMC. It contains
1,812 identities. Following the evaluation protocol \cite{zheng2017unlabeled},
702 identities are used as the training set and the remaining 1,110
identities as the testing set. During testing, one query image for
each identity in each camera is used for query and the remaining as
the gallery set. 

\noindent \noindent \textbf{CUHK01} \cite{cuhk01}  has 971 identities with 2
images per person captured in two disjoint camera views respectively. As in \cite{cuhk01},
we use as probe the images of camera A and utilize those from camera
B as gallery. 486 identities are randomly selected for testing and
the remaining are used for training. The experiments are repeated
for 10 times with the average results reported.

\begin{table}[t]
\centering{}%
\begin{tabular}{@{\extracolsep{\fill}}c||cc|cc}
\hline 
\multirow{2}{*}{{\small{}Methods}} & \multicolumn{2}{c|}{{\small{}{}{}Single-Query}} & \multicolumn{2}{c}{{\small{}{}{}Multi-Query}}\tabularnewline
\cline{2-5} 
 & {\small{}{}{}{}R-1}  & {\small{}{}{}{}mAP}  & {\small{}{}{}{}R-1}  & {\small{}{}{}{}mAP }\tabularnewline
\hline 
{\small{}{}TMA \cite{martinel2016eccv}}  & {\small{}{}{}{}47.90}  & {\small{}{}{}{}22.3}  & {\small{}{}{}{}\textendash{}}  & {\small{}{}{}{}\textendash{} }\tabularnewline
{\small{}{}{}{}SCSP \cite{chen2016cvpr}}  & {\small{}{}{}{}51.90}  & {\small{}{}{}{}26.40}  & {\small{}{}{}{}\textendash{}}  & {\small{}{}{}{}\textendash{} }\tabularnewline
{\small{}{}{}{}DNS \cite{null_space_cvpr2016}}  & {\small{}{}{}{}61.02}  & {\small{}{}{}{}35.68}  & {\small{}{}{}{}71.56}  & {\small{}{}{}{}46.03}\tabularnewline
{\small{}{}{}{}LSTM Siamese \cite{lstm2016eccv}}  & {\small{}{}{}{}\textendash{}}  & {\small{}{}{}{}\textendash{}}  & {\small{}{}{}{}61.60}  & {\small{}{}{}{}35.31}\tabularnewline
{\small{}{}{}{}Gated\_Sia \cite{gated_siamese_eccv2016}}  & {\small{}{}{}{}65.88}  & {\small{}{}{}{}39.55}  & {\small{}{}{}{}76.50}  & {\small{}{}{}{}48.50 }\tabularnewline
{\small{}{}{}{}HP-net \cite{liu2017hydraplus}}  & {\small{}{}{}{}76.90}  & {\small{}{}{}{}\textendash{}}  & {\small{}{}{}{}\textendash{}}  & {\small{}{}{}{}\textendash{}}\tabularnewline
{\small{}{}{}{}Spindle \cite{zhao2017spindle}}  & {\small{}{}{}{}76.90}  & {\small{}{}{}{}\textendash{}}  & {\small{}{}{}{}\textendash{}}  & {\small{}{}{}{}\textendash{}}\tabularnewline
{\small{}{}{}Basel.+LSRO \cite{zheng2017unlabeled}{*}}  & {\small{}{}{}{}78.06}  & {\small{}{}{}56.23}  & {\small{}{}{}85.12}  & {\small{}{}{}68.52}\tabularnewline
{\small{}{}{}{}PIE \cite{zheng2017pose}}  & {\small{}{}{}{}79.33}  & {\small{}{}{}{}55.95}  & {\small{}{}{}{}\textendash{}}  & {\small{}{}{}{}\textendash{} }\tabularnewline
{\small{}{}{}{}Verif.-Identif. \cite{verif}}  & {\small{}{}{}{}79.51}  & {\small{}{}{}{}59.87}  & {\small{}{}{}{}85.84}  & {\small{}{}{}{}70.33 }\tabularnewline
{\small{}{}{}{}DLPAR\cite{zhao2017deeply}}  & {\small{}{}{}{}81.00}  & {\small{}{}{}{}63.40}  & {\small{}{}{}{}\textendash{}}  & {\small{}{}{}{}\textendash{} }\tabularnewline
{\small{}{}{}{}DeepTransfer \cite{deeptransfer2016}}  & {\small{}{}{}{}83.70}  & {\small{}{}{}{}65.50}  & {\small{}{}{}{}89.60}  & {\small{}{}{}{}73.80 }\tabularnewline
{\small{}{}{}Verif-Identif.+LSRO{}\cite{zheng2017unlabeled}{*}}  & {\small{}{}{}{}83.97}  & {\small{}{}{}{}66.07}  & {\small{}{}{}{}88.42}  & {\small{}{}{}{}76.10 }\tabularnewline
{\small{}{}{}{}PDC \cite{su2017pose}}  & {\small{}{}{}{}84.14}  & {\small{}{}{}{}63.41}  & {\small{}{}{}{}\textendash{}}  & {\small{}{}{}{}\textendash{} }\tabularnewline
{\small{}{}{}{}DML \cite{zhang2017deep}}  & {\small{}{}{}{}87.7}  & {\small{}{}{}{}68.8}  & {\small{}{}{}{}\textendash{}}  & {\small{}{}{}{}\textendash{} }\tabularnewline
{\small{}{}{}{}SSM \cite{bai2017scalable}}  & {\small{}{}{}{}82.2}  & {\small{}{}{}{}68.8}  & {\small{}{}{}{}88.2}  & {\small{}{}{}{}76.2}\tabularnewline
{\small{}{}{}{}JLML \cite{li2017person}} & {\small{}{}{}{}85.10}  & {\small{}{}{}{}65.50}  & {\small{}{}{}{}89.70}  & {\small{}{}{}{}74.50}\tabularnewline
\hline 
\hline 
{\small{}{}ResNet-50-A}  & {\small{}{}{}{}87.26}  & {\small{}{}{}{}69.32}  & 91.81 & 77.85\tabularnewline
\hline 
{\small{}{}{}{}Ours (SL)} & \textbf{\small{}{}{}{}89.43}{\small{} } & \textbf{\small{}{}{}{}72.58{}}\textbf{ } & \textbf{92.93} & \textbf{80.19}\tabularnewline
\hline 
\end{tabular}\caption{\label{tab:Results-of-market}Results on Market-1501. `-' indicates
not reported. Note that {*}: {\small{}{}{}on \cite{zheng2017unlabeled},
we report the results of using both }Basel.+LSRO and Verif-Identif.+LSRO.
Our model only uses the identification loss, so should be compared
with Basel. + LSRO which uses the same ResNet-50 base network and
the same loss.}
\end{table}

\vspace{0.1in}

\noindent \textbf{Evaluation metrics}. Two evaluation metrics are
used to quantitatively measure the re-id performance. The first one
is Rank-1, Rank-5 and Rank-10 accuracy. For Market-1501 and DukeMTMC-reID
datasets, the mean Average Precision (mAP) is also used.

\noindent \textbf{Implementation details.} Our model is implemented
on Tensorflow \cite{tensorflow} (PN-GAN part) and Caffe \cite{caffe}
(re-id feature learning part) framework. The $\lambda_{1}$ 
in Eq (\ref{eq:generator}) is empirically set as 10
in all experiments. We utilize the two-stepped fine-tuning strategy
in \cite{geng2016deep} to fine-tune ResNet-50-A and ResNet-50-B.
The input images are resized into $256\times128$. Adam \cite{adam}
is used to train both the PN-GAN model and re-id networks with a learning
rate of 0.0002, $\beta_{1}=0.5$, a batch size of 32, and a learning
rate of 0.00035, $\beta_{1}=0.9$, a batch size of 16, respectively.
The dropout ratio is set as 0.5. Our PN-GAN models and re-id networks
are converged in 19 hours and 8 hours individually on Market-1501
with one NVIDIA 1080Ti GPU card.  Codes and trained models will
be made available on the first author's webpage.

\begin{table*}[t]
\begin{centering}
\begin{tabular}{cc}
\begin{tabular}{c||ccc}
\hline 
{\small{}{}{}Method}  & {\small{}{}{}R-1}  & {\small{}{}{}R-5}  & {\small{}{}{}R-10}\tabularnewline
\hline 
{\small{}{}DeepReid \cite{deepreid}}  & {\small{}{}{}19.89}  & {\small{}{}{}50.00}  & {\small{}{}{}64.00}\tabularnewline
{\small{}{}{}Imp-Deep \cite{Ejaz_cvpr2015}}  & {\small{}{}{}44.96}  & {\small{}{}{}76.01}  & {\small{}{}{}83.47}\tabularnewline
{\small{}{}{}EMD \cite{hailin_shi}}  & {\small{}{}{}52.09}  & {\small{}{}{}82.87}  & {\small{}{}{}91.78}\tabularnewline
{\small{}{}SI-CI \cite{joint_learning_cvpr16}}  & {\small{}{}{}52.17}  & {\small{}{}{}84.30}  & {\small{}{}{}92.30}\tabularnewline
{\small{}{}{}LSTM Siamese \cite{lstm2016eccv}}  & {\small{}{}{}57.30}  & {\small{}{}{}80.10}  & {\small{}{}{}88.30}\tabularnewline
{\small{}{}{}PIE \cite{zheng2017pose}}  & {\small{}{}{}67.10}  & {\small{}{}{}92.20}  & {\small{}{}{}96.60}\tabularnewline
{\small{}{}{}Gated\_Sia \cite{gated_siamese_eccv2016}}  & {\small{}{}{}68.10}  & {\small{}{}{}88.10}  & {\small{}{}{}94.60}\tabularnewline
{\small{}{}{}Basel. + LSRO \cite{zheng2017unlabeled}{}}  & {\small{}{}{}73.10}  & {\small{}{}{}92.70}  & {\small{}{}{}96.70 }\tabularnewline
{\small{}{}{}DGD \cite{xiao2016learning}}  & {\small{}{}{}75.30}  & {\small{}{}{}\textendash{}}  & {\small{}{}{}\textendash{}}\tabularnewline
{\small{}{}{}OIM \cite{xiao2017joint}}  & {\small{}{}{}77.50}  & {\small{}{}{}\textendash{}}  & {\small{}{}{}\textendash{}}\tabularnewline
{\small{}{}{}PDC \cite{su2017pose}}  & {\small{}{}{}78.92}  & {\small{}{}{}94.83}  & {\small{}{}{}97.15}\tabularnewline
%{\small{}{}{}JLML \cite{li2017person}}  & {\small{}{}{}80.60}  & {\small{}{}{}\textendash{}}  & {\small{}{}{}\textendash{}}\tabularnewline
{\small{}{}{}DLPAR\cite{zhao2017deeply}}  & {\small{}{}{}}\textbf{\small{}81.60}  & \textbf{\small{}{}{}97.30}\textbf{ } & {\small{}{}{}98.40}\tabularnewline
\hline 
\hline 
{\small{}{}ResNet-50-A} (SL) & {\small{} 76.83}  & {\small{}{}{}{} 93.79}  & {\small{}{}{}{}97.27}\tabularnewline
{\small{}{}{}Ours (SL)}  & {\small{}79.76 } & {\small{}{}{}{}96.24 } & \textbf{\small{}{}{}{}98.56}\tabularnewline
\hline 
\hline 
{\small{}ResNet-50-A (TL)} & 16.50  & 38.60 & 52.84\tabularnewline
Ours (TL) & 16.85 & 39.05 & 53.32\tabularnewline
\hline 
\end{tabular} & %
\begin{tabular}{c||ccc}
\hline 
{\small{}{}{}Method}  & {\small{}{}{}R-1}  & {\small{}{}{}R-5}  & {\small{}{}{}R-10}\tabularnewline
\hline 
{\small{}{}{}eSDC \cite{unsupervised_per_reid}}  & {\small{}{}{}19.76}  & {\small{}{}{}32.72}  & {\small{}{}{}40.29}\tabularnewline
{\small{}{}{}kLFDA \cite{kLFDA}}  & {\small{}{}{}32.76}  & {\small{}{}{}59.01}  & {\small{}{}{}69.63}\tabularnewline
{\small{}{}{}mFilter \cite{mFilter}}  & {\small{}{}{}34.30}  & {\small{}{}{}55.00}  & {\small{}{}{}65.30}\tabularnewline
{\small{}{}{}Imp-Deep \cite{Ejaz_cvpr2015}}  & {\small{}{}{}47.53}  & {\small{}{}{}71.50}  & {\small{}{}{}80.00}\tabularnewline
{\small{}{}{}DeepRanking \cite{deepranking2016TIP}}  & {\small{}{}{}50.41}  & {\small{}{}{}75.93}  & {\small{}{}{}84.07}\tabularnewline
{\small{}{}{}Ensembles \cite{Ensembles}}  & {\small{}{}{}53.40}  & {\small{}{}{}76.30}  & {\small{}{}{}84.40}\tabularnewline
{\small{}{}{}ImpTrpLoss \cite{ImpTrpLoss}}  & {\small{}{}{}53.70}  & {\small{}{}{}84.30}  & {\small{}{}{}91.00}\tabularnewline
{\small{}{}{}GOG \cite{GOG}}  & {\small{}{}{}57.80}  & {\small{}{}{}79.10}  & {\small{}{}{}86.20}\tabularnewline
{\small{}{}{}Quadruplet \cite{quadruplet}}  & {\small{}{}{}62.55}  & {\small{}{}{}83.44}  & {\small{}{}{}89.71}\tabularnewline
{\small{}{}{}NullReid \cite{NullReid}}  & {\small{}{}{}64.98}  & {\small{}{}{}84.96}  & {\small{}{}{}89.92}\tabularnewline
\hline 
\hline 
{\small{}{}ResNet-50-A} (SL) & {\small{}{}{}64.56}  & {\small{}{}{}83.66}  & {\small{}{}{}89.74}\tabularnewline
{\small{}{}{}Ours (SL)}  & \textbf{\small{}{}{}67.65}{\small{} } & \textbf{\small{}{}{}86.64}{\small{} } & \textbf{\small{}{}{}91.82}\tabularnewline
\hline 
{\small{}{}ResNet-50-A} (TL) & 27.20 & 48.60 & 59.20\tabularnewline
Ours (TL)  & 27.58 & 49.17 & 59.57\tabularnewline
\hline 
\end{tabular}\tabularnewline
(a) Results on CUHK03  & (b) Results on CUHK01\tabularnewline
\end{tabular}
\par\end{centering}
\caption{\label{tab:Results-of-CUHK03.}Results on CUHK01 and CUHK03 datasets.
Note that both Spindle \cite{zhao2017spindle} and HP-net \cite{liu2017hydraplus}
reported higher results on CUHK03. But their results are obtained
using a very different setting: six auxiliary re-id datasets are used
and both labeled and detected bounding boxes are used for both training
and testing. So their results are not comparable to those in this
table.}
\end{table*}

\noindent \textbf{Experimental Settings.} Experiments are conducted
under two settings. The first is the standard \textbf{Supervised Learning}
(SL) setting on all datasets: the models are trained on the training
set of the dataset, and evaluated on the testing set. The other one
is the \textbf{Transfer Learning} (TL) setting only for the datasets,
CUHK03, CUHK01, and DukeMTMC-reID. Specifically, the re-id model is
trained on Market-1501 dataset. We then directly utilize the trained
single model to do the testing (\emph{i.e}., to synthesize images
with canonical poses and to extract the nine feature vectors) on the
test set of CUHK03, CUHK01, and DukeMTMC-reID. That is, no model updating
is done using any data from these three datasets. The TL setting is
especially useful in real-world scenarios, where a pre-trained model
needs to be deployed to a new camera network without any model fine-tuning.
This setting thus tests how generalizable a re-id model is.

\subsection{Supervised Learning Results}

\begin{table}
\begin{centering}
{\small{}}%
\begin{tabular}{c||ccc}
\hline 
{\small{}{}Methods } & {\small{}{}{}R-1 } & {\small{}{}{}R-10 } & {\small{}{}{}mAP }\tabularnewline
\hline 
{\small{}LOMO+XQDA\cite{XQDA} } & {\small{}{}30.80 } & {\small{}{}\textendash{} } & {\small{}{}17.00}\tabularnewline
{\small{}{}ResNet50 \cite{resnet} } & {\small{}{}65.20 } & {\small{}{}\textendash{} } & {\small{}{}45.00}\tabularnewline
{\small{}{}Basel. +LSRO \cite{zheng2017unlabeled} } & {\small{}{}67.70 } & {\small{}{}\textendash{} } & {\small{}{}47.10}\tabularnewline
{\small{}{}AttIDNet \cite{lin2017improving} } & {\small{}{}70.69 } & {\small{}{}\textendash{} } & {\small{}{}51.88}\tabularnewline
%{\small{}{}SVDNet \cite{SVDNet} } & {\small{}{}}\textbf{\small{}76.70 } & \textbf{\small{}{}89.90 } & \textbf{\small{}{}56.80}\tabularnewline
\hline 
\hline 
{\small{}ResNet-50-A (SL)} & 72.80 & 87.90 & 52.48\tabularnewline
{\small{}{}{}Ours (SL)} & \textbf{\small{}{}73.58} & \textbf{\small{}{}88.75} & \textbf{\small{}{}53.20}\tabularnewline
\hline 
{\small{}ResNet-50-A (TL)} & 27.872 & 51.122 & 13.942\tabularnewline
{\small{}{}{}Ours (TL)} & 29.937 & 51.615 & 15.768\tabularnewline
\hline 
\end{tabular}
\par\end{centering}{\small \par}
\caption{\label{tab:Results-on-the-Duke}Results on DukeMTMC-reID.}
\end{table}

\noindent \textbf{Results on large-scale datasets}. Tables \ref{tab:Results-of-market},
\ref{tab:Results-on-the-Duke} and \ref{tab:Results-of-CUHK03.} (a)
compare our model with the best performing alternative models. We
can make the following observations:

\noindent (1) On all three datasets, the results clearly show that,
in the supervised learning settings, our results are improved over
those of ResNet-50-A baselines by a clear margin. This validates that
the synthetic person images generated by PN-GAN can indeed help the
person re-id tasks. %Notably on Market-1501, even with our ResNet-50-A model alone
%can obtain $87.34\%$ Rank-1 accuracy, as well as $69.32$ mAP, which
%are higher than those of most of the other baselines. This shows the
%ResNet-50 is a very good and strong baseline in person re-id tasks.

\noindent \noindent (2) Compared with the existing pose-guided re-id
models \cite{zhao2017spindle,zheng2017pose,su2017pose}, our model
is clearly better, indicating that synthesizing multiple normalized
poses is a more effective way to deal with the large pose variation
problem. 

\noindent \noindent (3) Compared with the only other re-id model
that uses synthesized images for re-id model training \cite{zheng2017unlabeled},
our model yields better performance for all datasets, the gap on Market-1501
and DukeMTCM-reID being particularly clear. This is because our model
can synthesize images with different poses, which can thus be used
for supervised training. In contrast, the synthesized images in \cite{zheng2017unlabeled}
do not correspond to any particular person identities or poses, so
can only be used as unlabeled or weakly-labeled data. 

\noindent \textbf{Results on small-scale dataset}. On the smaller
dataset \textendash{} CUHK01, Table \ref{tab:Results-of-CUHK03.}(b)
shows that, again our ResNet-50-A is a pretty strong baseline which
can beat almost all the other methods. And by using the normalized
pose images generated by PN-GAN, our framework further boosts the
performance of ResNet-50-A by more than $3\%$ in the supervised setting.
This demonstrates the efficacy of our framework. Note that on the
small dataset CUHK01, the handcrafted feature + metric learning based
models (e.g., NullReid \cite{NullReid}) are still
quite competitive, often beating the more recent deep models. This
reveals the limitations of the existing deep models on scalability
and generalizability. In particular, previous deep re-id models are
pre-trained on some large-scale training datasets, such as CUHK03
and Market-1501. But the models still struggle to fine-tune on the
small datasets such as CUHK01 due to the covariate condition differences
between them. With the pose normalization, our model is more adaptive
to the small datasets and the model pre-trained on only Market-1501
can be easily fine-tuned on the small datasets, achieving much better
result than existing models.

\subsection{Transfer Learning Results}

\begin{table*}
\begin{centering}
\begin{tabular}{c|cc|cc|cc|cc}
\hline 
{\small{}{}Dataset}  & \multicolumn{2}{c|}{{\small{}{}Market-1501}} & \multicolumn{2}{c|}{{\small{}{}DukeMTMC-reID}} & \multicolumn{2}{c|}{{\small{}{}CUHK03}} & \multicolumn{2}{c}{{\small{}{}CUHK01}}\tabularnewline
\hline 
{\small{}{}Methods}  & {\small{}{}R-1}  & {\small{}{}mAP}  & {\small{}{}R-1}  & {\small{}{}mAP}  & {\small{}{}R-1}  & {\small{}{}R-5}  & {\small{}{}R-1}  & {\small{}{}R-5} \tabularnewline
\hline 
\hline 
{\small{}{}ResNet-50-A}  & {\small{}{}{}87.26}  & {\small{}{}{}69.32}  & {\small{}{}72.80}  & {\small{}{}52.48}  & {\small{}{}76.83} & {\small{}{}93.79}  & {\small{}{}64.56}  & {\small{}{}83.66} \tabularnewline
{\small{}{}ResNet-50-B}  & {\small{}{}{}63.75}  & {\small{}{}{}41.29}  & {\small{}{}26.62}  & {\small{}{}14.30}  & {\small{}{}32.54}  & {\small{}{}55.12}  & {\small{}{}36.18}  & {\small{}{}51.17} \tabularnewline
\hline 
{\small{}{}Ours}  & \textbf{\small{}{}{}89.43}{\small{}{}}  & \textbf{\small{}{}{}72.58}{\small{} } & \textbf{\small{}{}73.58}{\small{} } & \textbf{\small{}{}53.20}{\small{} } & \textbf{\small{}{}79.76 }{\small{}{}}  & \textbf{\small{}{}96.24}{\small{}{}}  & \textbf{\small{}{}67.65}{\small{}{}}  & \textbf{\small{}{}86.64}{\small{}{}} \tabularnewline
\hline 
\end{tabular}
\par\end{centering}
\caption{\label{tab:The-Ablation-Study-1}The Ablation Study of Rank-1 and
Rank-5 on benchmarks. }
\end{table*}

\begin{table}
\begin{centering}
{\small{}}%
\begin{tabular}{@{\extracolsep{\fill}}c|cc|cc}
\hline 
{\small{}Feature(s) } & \multicolumn{2}{c|}{{\small{}{}{}1 pose}} & \multicolumn{2}{c}{{\small{}8 poses}}\tabularnewline
\hline 
{\small{}Methods } & {\small{}R-1 } & {\small{}mAP } & {\small{}{}{}R-1 } & {\small{}mAP}\tabularnewline
\hline 
\hline 
{\small{}ResNet-50-A } & {\small{}87.26} & {\small{}69.34} & {\small{}87.26} & {\small{}69.34}\tabularnewline
{\small{}ResNet-50-B } & 58.70 & 36.69 & {\small{}63.75} & {\small{}41.67}\tabularnewline
\hline 
{\small{}Ours (SL) } & 87.65 & 69.60 & {\small{}89.40} & {\small{}72.58}\tabularnewline
\hline 
\end{tabular}
\par\end{centering}{\small \par}
\caption{\label{tab:The-Ablation-Study-2}The Ablation Study of Market-1501
on 1 pose feature and 8 pose features. }
\end{table}

We report our results obtained under the TL settings on the three
datasets \textendash{} CUHK03, CUHK01, and DukeMTMC-reID in Table
\ref{tab:Results-of-CUHK03.}(b), and Table \ref{tab:Results-on-the-Duke}
respectively. On\textbf{ }CUHK01 dataset, we can achieve $27.58\%$
Rank-1 accuracy in Table \ref{tab:Results-of-CUHK03.}(b) which is
comparable to some models trained under the supervised learning setting,
such as eSDC {\cite{unsupervised_per_reid}}. These results
thus show that our model has the potential to be truly generalizable
to a new re-id data from new camera networks \textendash{} when operating
in a `plug-and-play' mode. Our results are also compared against those
of ResNet-50-A (TL) baseline. On all three datasets, we can observe that
our model gets improved over those of ResNet-50-A (TL) baseline. Again,
this demonstrates that our pose normalized person images can also
help the person re-id in the transfer learning settings. Note that
due to the intrinsic difficulty of transfer setting, the results are
still much lower than those in supervised setting. 

\subsection{Further Evaluations}

\noindent \textbf{Ablation Studies}. We first evaluate the contributions
from the two types of features computed using ResNet-50-A and ResNet-50-B
respectively towards the final performance. Table \ref{tab:The-Ablation-Study-2}
shows that: (1) Each model on its own is quite strong - better than
many existing models compared earlier.  (2) When the two types of features
are combined, there is an improvement in the final results on all four
datasets. This clearly indicates that the two types of features are
complementary to each other. In a second study, we compare the result
obtained when features are merged with 8 poses and that obtained with
only one pose, in Table \ref{tab:The-Ablation-Study-2}. The result
drops from $72.58$ to $69.60$ on Market-1501 on mAP. This suggests
that having eight canonical poses is beneficial \textendash{} the
quality of generated image under one particular pose may be poor;
using all eight poses thus reduces the sensitivity to the quality
of the generated images for specific poses.

\begin{figure*}
\begin{centering}
\includegraphics[scale=0.2]{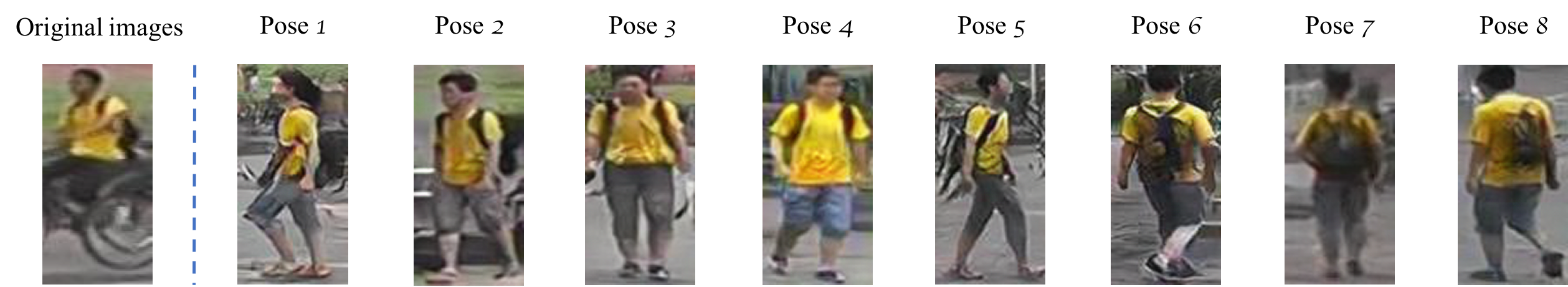}\caption{\label{fig:Visualization}Visualization of different poses generated
by PN-GAN model.}
\par\end{centering}
\end{figure*}

\noindent \textbf{Examples of the synthesized images}. Figure \ref{fig:Visualization}
gives some examples of the synthesized image poses. Given one input
image, our image generator can produce realistic images under different
poses, while keeping the similar visual appearance as the input person
image. We find that, (1) Even though we did not explicitly use
the attributes to guide the PN-GAN,  the generated images of different
poses have roughly the same visual attributes as the original
images. (2) Our model can help alleviate the problems caused by occlusion
as shown in the last row of Fig.~\ref{fig:Visualization}: a man with
yellow shirt and grey trousers is blocked by a bicycle, while our image
generator can generate synthesized images to keep his key attributes
whilst removing the occlusion. 

\section{Conclusion}

We have proposed a novel deep person image generation model by synthesizing
pose-normalized person images for re-id. In contrast to previous re-id
approaches that try to extract discriminator features which are identity-sensitive
but view-insensitive, the proposed method learns complementary features
from both original images and pose-normalized synthetic images. Extensive
experiments on four benchmarks showed that our model achieves state-of-the-art performance. More importantly, we demonstrated that our model
can be generalized to new re-id datasets collected from new camera
networks without any additional data collection and model fine-tuning.

{\small{}\bibliographystyle{ieee}
\bibliography{egbib}
 }{\small \par}
\end{document}